\documentclass[10pt, a4paper]{article}

\usepackage[final]{lrec2026} 

\usepackage{multirow}
\usepackage{adjustbox}
\usepackage{arydshln}
\usepackage{booktabs} 
\usepackage{verbatim} 

\usepackage{soul}
\soulregister{\ref}{1}
\sethlcolor{yellow}

\newcommand{\subsetting}[1]{\hspace{1.2em}\textit{#1}}

\title{MzansiText and MzansiLM: An Open Corpus and Decoder-Only Language Model for South African Languages}

\name{\parbox{0.95\textwidth}{\centering
Anri Lombard, Simbarashe Mawere, Temi Aina, Ethan Wolff, Sbonelo Gumede, Elan Novick, Francois Meyer, and Jan Buys}}

\address{University of Cape Town \\
         Cape Town, South Africa \\
         \{LMBANR001, MWRSIM003, ANXTEM001, WLFETH001, GMDSBO006, NVCELA001\}@myuct.ac.za \\
         \{francois.meyer, jan.buys\}@uct.ac.za\\}

\abstract{
Decoder-only language models can be adapted to diverse tasks through instruction finetuning, but the extent to which this generalizes at small scale for low-resource languages remains unclear. We focus on the languages of South Africa, where we are not aware of a publicly available decoder-only model that explicitly targets all eleven official written languages, nine of which are low-resource. We introduce MzansiText, a curated multilingual pretraining corpus with a reproducible filtering pipeline, and MzansiLM, a 125M-parameter language model trained from scratch. We evaluate MzansiLM on natural language understanding and generation using three adaptation regimes: monolingual task-specific finetuning, multilingual task-specific finetuning, and general multi-task instruction finetuning. Monolingual task-specific finetuning achieves strong performance on data-to-text generation, reaching 20.65 BLEU on isiXhosa and competing with encoder-decoder baselines over ten times larger. Multilingual task-specific finetuning benefits closely related languages on topic classification, achieving 78.5\% macro-F1 on isiXhosa news classification. While MzansiLM adapts effectively to supervised NLU and NLG tasks, few-shot reasoning remains challenging at this model size, with performance near chance even for much larger decoder-only models. We release MzansiText and MzansiLM to provide a reproducible decoder-only baseline and clear guidance on adaptation strategies for South African languages at small scale.
\\ \newline 
}

\begin{document}

\maketitleabstract

\section{Introduction}

The application of large language models (LLMs) to natural language processing tasks has progressed through three phases: pretrained models finetuned for specific tasks (BERT, T5) \citep{devlin-etal-2019-bert,raffel2020exploring}, in-context learning at inference time (GPT-3) \citep{brown2020language}, and instruction finetuning for broader generalization \citep{wei2022finetuned,wang2023selfinstruct}. Each phase has relied on increasingly larger models and more pretraining data, limiting direct applicability to low-resource settings.

For African languages, language model research has historically been led by encoder-only and encoder--decoder approaches \citep{ogueji-etal-2021-small,alabi-etal-2022-adapting,jude-ogundepo-etal-2022-afriteva,adelani-etal-2022-thousand}, although decoder-only work has recently increased \citep{tonja2024inkubalm,uemura-etal-2024-afriinstruct,buzaaba2025lugha-llama}. Even so, decoder-only research for African languages remains comparatively early-stage, and many of the strongest established baselines are still encoder-only or encoder--decoder models. In this paper, we focus on the languages of South Africa, a country with particularly high linguistic diversity.

South Africa recognizes eleven official written languages: Sepedi (Northern Sotho), Sesotho, Setswana, siSwati, Tshivenda, Xitsonga, Afrikaans, English, isiNdebele, isiXhosa, and isiZulu. Afrikaans and English are Indo-European languages in the West Germanic branch, while the remaining nine are Bantu languages. 
Bantu languages are agglutinative and exhibit rich noun-class systems \citep{nurse2004bantu}. Table~\ref{tab:sa_lang_context} summarizes home-language usage and shows that only 8.7\% of South Africans speak English at home \citep{statssa2024census}, which underscores the value of language technologies that work across the full set of official languages. 

This paper introduces MzansiText and MzansiLM, a curated multilingual pretraining corpus and a 125M-parameter decoder-only model trained from scratch with pretraining coverage over all eleven official written South African languages. The dataset is available on HuggingFace at \href{https://huggingface.co/datasets/anrilombard/mzansi-text}{\tt anrilombard/mzansi-text} and the model at \href{https://huggingface.co/anrilombard/mzansilm-125m}{\tt anrilombard/mzansilm-125m}.

We systematically evaluate three adaptation regimes: task-specific monolingual finetuning, task-specific multilingual finetuning, and general multi-task instruction finetuning. MzansiLM achieves competitive performance on some generation and sequence labelling tasks despite its modest parameter count, in some cases surpassing decoder-only models over ten times larger, though encoder-based architectures retain advantages on classification and other structured prediction tasks and few-shot reasoning remains challenging at this scale.

To summarise, this paper contributes:
\begin{itemize}
\item MzansiText, a curated pretraining corpus with a reproducible preparation pipeline and pretraining coverage over all eleven official written South African languages.
\item MzansiLM, a 125M-parameter decoder-only model trained from scratch on MzansiText.
\item A systematic comparison of different instruction finetuning approaches across available tasks for South African languages, revealing where decoder-only models excel and where they still fall short.
\end{itemize}
We also publicly release our code\footnote{\href{https://github.com/Anri-Lombard/sallm}{\tt github.com/Anri-Lombard/sallm}} for data preparation, pretraining, and evaluation.

\begin{table}[t]
\centering
\small
\setlength{\tabcolsep}{5pt}
\begin{tabular}{l l r r}
\hline
\textbf{Language} & \textbf{ISO} & \textbf{Speakers} & \textbf{\%} \\
\hline
isiZulu & zul & 14{,}613{,}202 & 24.4 \\
isiXhosa & xho & 9{,}786{,}928 & 16.3 \\
Afrikaans & afr & 6{,}365{,}488 & 10.6 \\
Sepedi (Northern Sotho) & nso & 5{,}972{,}255 & 10.0 \\
English & eng & 5{,}228{,}301 & 8.7 \\
Setswana & tsn & 4{,}972{,}787 & 8.3 \\
Sesotho & sot & 4{,}678{,}964 & 7.8 \\
Xitsonga & tso & 2{,}784{,}279 & 4.7 \\
siSwati & ssw & 1{,}692{,}719 & 2.8 \\
Tshivenda & ven & 1{,}480{,}565 & 2.5 \\
isiNdebele & nbl & 1{,}044{,}377 & 1.7 \\
\hline
\end{tabular}
\caption{The number of L1 speakers for South Africa's eleven official written languages \citep{statssa2024census}. 
\label{tab:sa_lang_context}
}
\end{table}

\section{Related Work}


Current research in African NLP increasingly explores decoder-only architectures, though strong open-weight baselines are still predominantly encoder-only and encoder--decoder models. AfriBERTa \citep{ogueji-etal-2021-small} provides encoder-only baselines trained from scratch for several African languages. AfroXLMR \citep{alabi-etal-2022-adapting} adapts XLM-R \citep{conneau-etal-2020-unsupervised} to African languages through continued pretraining.
On the encoder--decoder side, AfriTeVa \citep{jude-ogundepo-etal-2022-afriteva,oladipo-etal-2023-wura} builds sequence-to-sequence models by pretraining on the WURA corpus, which improves quality for African languages. 
Afri-mT5 \citep{adelani-etal-2022-thousand} adapts mT5 \citep{xue-etal-2021-mt5} for African languages through continued pretraining. Aya-101 \citep{ustun-etal-2024-aya} is an instruction-tuned encoder--decoder model covering 101 languages, providing a broad multilingual reference point.

A small but growing body of work investigates decoder-only models for African languages. InkubaLM \citep{tonja2024inkubalm} is a 0.4B parameter LM trained from scratch on a corpus that includes isiZulu, isiXhosa, Yoruba, Hausa, and Swahili,
achieving competitive results on AfriXNLI and AfriMMLU, outperforming SmolLM-1.7B \citep{allal2024smollm} and matching or exceeding LLaMA 3-8B \citep{dubey2024llama3} on these tasks despite having far fewer parameters. 
AfroLlama-V1 \citetlanguageresource{health2024afrollama} applies continual pretraining and instruction tuning to Llama-3-8B for Swahili, isiXhosa, isiZulu, Yoruba, Hausa, and English. AfriInstruct \citep{uemura-etal-2024-afriinstruct} continues pretraining LLaMA-2-7B on African languages, subsequently instruction tuning on diverse African tasks. Lugha-Llama \citep{buzaaba2025lugha-llama} adapts Llama-3.1-8B \citep{buzaaba2025lugha-llama} through continued pretraining for African languages.
Our work addresses three gaps. First, prior decoder-only models cover only subsets of South African languages (mainly Afrikaans, isiXhosa, isiZulu), never all eleven. Second, adaptation strategies are seldom compared: some focus on continued pretraining \citep{alabi-etal-2022-adapting,jude-ogundepo-etal-2022-afriteva,buzaaba2025lugha-llama}, others on instruction tuning \citep{uemura-etal-2024-afriinstruct}, but systematic comparisons across task-specific monolingual, multilingual, and multi-task instruction finetuning remain absent. Third, evaluation centers on broad benchmarks that often exclude languages at the lower end of the resource-availability spectrum
(IrokoBench \citep{adelani-etal-2025-irokobench}, AfriQA \citep{ogundepo-etal-2023-cross}) rather than comprehensive South African language–specific datasets covering both understanding and generation.

\section{Pretraining Dataset (MzansiText)}
\label{sec:dataset}

Training a decoder-only language model from scratch first requires a suitable pretraining corpus. We therefore construct MzansiText, a curated multilingual corpus for the eleven official written South African languages, by aggregating public sources and applying a reproducible preparation pipeline based on established best practices for large-scale web and multilingual corpora. This section introduces the data sources, preparation pipeline, and resulting token distribution.

\paragraph{Sources}
We draw on public language resources with substantive coverage or relevance to South African languages.

\begin{itemize}
  \item mC4 \citep{raffel2020exploring} consists of Common Crawl web text across 101 languages with filtering and deduplication. 
  
  \item CulturaX \citeplanguageresource{culturax} is a large multilingual dataset that applies a multi-stage language identification and cleaning process. 
  
  \item WURA \citeplanguageresource{wura} cleans mC4 for African languages and combines it with targeted crawls of local news sites. 
  
  \item Glot500-c \citeplanguageresource{glot500} covers 511 languages, aggregating multiple sources and applying cleaning and deduplication.
  
  \item NCHLT Text  \citeplanguageresource{nchlt} provides curated monolingual corpora for South African languages developed through national initiatives. The material is smaller, but cleaner, than other web crawled corpora. 
  
  \item CC100 \citep{wenzek-etal-2020-ccnet} comprises monolingual data extracted from Common Crawl. It provides broad coverage but inconsistent quality, so further filtering is applied for our targets.
  
  \item ParaCrawl \citeplanguageresource{paracrawl} is sentence-aligned parallel text mined from the web. It is useful as an indicator of in-language content and terminology, although alignment noise and domain skew are known challenges and are mitigated with strict filtering and deduplication. We use the monolingual African languages data drived from ParaCrawl for WMT2022. 
  
  \item Inkuba-Mono  \citeplanguageresource{inkuba-mono} provides monolingual corpora used for the Inkuba family of LMs \citep{tonja-etal-2024-inkubalm}. It includes isiZulu and isiXhosa among five African languages, and is based on public web text and news sources for specific languages.
\end{itemize}

\paragraph{Language balance}
We split the pretraining data into 80\%/10\%/10\% train, validation, and test splits. However, to prevent high-resource languages from dominating early stopping and model selection, the validation and test sets are capped at two million tokens per language. Unless otherwise stated, all token counts are computed with the 65{,}536-token tokenizer described in Section~\ref{sec:pretrain}. Table~\ref{tab:pretrain_distribution} reports per-language token counts and within-split shares. Note that we use English data only from WURA, Glot500, and NCHLT Text.

Due to the resource imbalance between languages, the resulting distribution is heavily skewed toward Afrikaans and English (approximately 84\% combined). We experimented with a more balanced pretraining mixture by capping English and Afrikaans to 25\% each in the training data. However, this balanced version led to measurable degradation on downstream tasks (see Appendix \ref{app:balanced_results}). We therefore retained the natural distribution to maximise usable high-quality data volume in this low-resource setting, while still ensuring pretraining coverage for every official language.

\paragraph{Preparation pipeline}
We implement a deterministic multi-stage pipeline with \texttt{Datatrove} to standardise quality across sources and languages \citep{datatrove-github}. The design follows best practices reported for large web corpora such as C4/T5 \citep{raffel2020exploring}, Gopher \citep{rae-etal-2021-gopher}, CulturaX \citep{nguyen-etal-2024-culturax}, and FineWeb \citep{penedo-etal-2024-fineweb}:
\begin{enumerate}
  \item Apply language identification at the document and line level, retaining segments whose dominant language is one of the eleven targets.
  \item Normalise and remove boilerplate, including HTML stripping, Unicode normalisation, control characters, and excess whitespace.
  \item Enforce structural constraints on length, script, and flag unexpected character distributions.
  \item Remove exact and near-duplicate content using hashing and MinHash-style signatures, both within and across sources.
  \item Apply safety screening to filter personally identifiable information and other sensitive content using conservative heuristics.
  \item Apply heuristic quality filters to remove templated, boilerplate, and low-information text.
  \item Pack short high-quality fragments into training records of roughly one thousand tokens while preserving sentence boundaries.
\end{enumerate}

Table~\ref{tab:source_stage_counts} reports token counts before and after deduplication and filtering. 
Across languages, deduplication removes 4.0 percent of the original word count, while subsequent filtering removes 22.9 percent of the deduplicated text, with the largest absolute reductions observed for Afrikaans, isiZulu, and isiXhosa.

\begin{table*}[t]
\centering
\small
\renewcommand{\arraystretch}{1.1}
\setlength{\tabcolsep}{8pt}
\begin{adjustbox}{center,max width=\textwidth}
\begin{tabular}{l | r r | r r | r r}
\hline
\multirow{2}{*}{\textbf{Language}} & \multicolumn{2}{c|}{\textbf{Train}} & \multicolumn{2}{c|}{\textbf{Validation}} & \multicolumn{2}{c}{\textbf{Test}} \\
 & \textbf{Tokens} & \textbf{\%} & \textbf{Tokens} & \textbf{\%} & \textbf{Tokens} & \textbf{\%} \\
\hline
Afrikaans & 2{,}475{,}913{,}822 & 64.96 & 1{,}865{,}255 & 14.42 & 1{,}875{,}605 & 14.24 \\
English & 740{,}994{,}679 & 19.44 & 1{,}813{,}651 & 14.02 & 1{,}821{,}803 & 13.83 \\
isiNdebele & 818{,}549 & 0.02 & 106{,}224 & 0.82 & 143{,}458 & 1.09 \\
Sepedi (Northern Sotho) & 6{,}697{,}358 & 0.18 & 685{,}425 & 5.30 & 778{,}656 & 5.91 \\
Sesotho & 97{,}558{,}939 & 2.56 & 2{,}315{,}298 & 17.90 & 2{,}316{,}170 & 17.59 \\
siSwati & 1{,}932{,}989 & 0.05 & 196{,}247 & 1.52 & 225{,}810 & 1.71 \\
Setswana & 10{,}082{,}930 & 0.26 & 1{,}216{,}539 & 9.41 & 1{,}413{,}473 & 10.73 \\
Xitsonga & 3{,}013{,}408 & 0.08 & 510{,}463 & 3.95 & 319{,}496 & 2.43 \\
Tshivenda & 1{,}852{,}481 & 0.05 & 191{,}495 & 1.48 & 243{,}315 & 1.85 \\
isiXhosa & 152{,}212{,}403 & 3.99 & 2{,}016{,}503 & 15.59 & 2{,}012{,}000 & 15.28 \\
isiZulu & 320{,}224{,}015 & 8.40 & 2{,}017{,}406 & 15.60 & 2{,}021{,}343 & 15.35 \\
\hline
Total & 3{,}811{,}301{,}573 & 100.00 & 12{,}934{,}506 & 100.00 & 13{,}171{,}129 & 100.00 \\
\hline
\end{tabular}
\end{adjustbox}
\caption{Token counts by language and split after language identification and filtering using our 65{,}536-token tokenizer. Percentages show each language’s share within a split. Validation and test apply a two-million-token per-language cap.}
\label{tab:pretrain_distribution}
\end{table*}

\begin{table*}[t]
\centering
\small
\renewcommand{\arraystretch}{1.1}
\setlength{\tabcolsep}{8pt}
\begin{tabular}{l r r r c}
\hline
\textbf{Source} & \textbf{Before processing} & \textbf{After deduplication} & \textbf{After filtering} & \textbf{Percent retained} \\
\hline
WURA      & 997{,}742{,}420  & 988{,}157{,}747  & 879{,}523{,}389  & 88.2 \\
mC4       & 1{,}008{,}467{,}039 & 979{,}623{,}283  & 824{,}839{,}355  & 81.8 \\
CulturaX  & 702{,}050{,}710   & 695{,}083{,}559  & 676{,}035{,}198  & 96.3 \\
Glot500   &  191{,}154{,}885  & 167{,}600{,}993  & 79{,}244{,}672   & 47.3 \\
Inkuba-Mono    & 234{,}941{,}750   & 196{,}457{,}594  & 63{,}114{,}818   & 26.9 \\
CC100     & 23{,}822{,}691    & 20{,}291{,}392   & 16{,}922{,}824   & 71.1 \\
ParaCrawl & 287{,}212{,}175   & 262{,}079{,}888  & 10{,}139{,}616   & 3.5 \\
NCHLT Text  & 13{,}473{,}354    & 11{,}372{,}194   & 9{,}840{,}573    & 73.0 \\
\hline
\end{tabular}
\caption{Per-source word counts at three stages of the pipeline, aggregated across languages.}
\label{tab:source_stage_counts}
\end{table*}

\section{Model Pretraining}
\label{sec:pretrain}

MzansiLM is a compact decoder-only model trained on MzansiText. We use the architecture configuration from MobileLLM at 125M parameters \citep{mobilellm}, which employs hyperparameters optimized for small-scale models through systematic ablation studies, and adopt a LLaMA-style decoder stack with causal self-attention \citep{touvron2023llama}. The maximum context length is 2{,}048. We train a BPE tokenizer \citep{sennrich-etal-2016-neural} with a vocabulary size of 65{,}536 on our pretraining split (all dataset token counts reported in this paper are computed with this tokenizer).

Optimisation and scheduling follow the MobileLLM recipe at the 125M scale. We train for five epochs over the training split and monitor cross-entropy loss and mean token accuracy on the training and validation sets. To avoid higher-resource languages from dominating the held-out loss, we average held-out losses across the languages. 
All runs use four NVIDIA A100 accelerators with 80\,GB of memory each. End-to-end pretraining completes in approximately 27 hours.

\begin{table*}[ht]
\centering
\small
\setlength{\tabcolsep}{4pt}
\begin{adjustbox}{center,max width=\textwidth}
\begin{tabular}{l l | r r | r r | r r}
\hline
\textbf{Task} & \textbf{Language} & \multicolumn{2}{c|}{\textbf{Train}} & \multicolumn{2}{c|}{\textbf{Validation}} & \multicolumn{2}{c}{\textbf{Test}} \\
\cline{3-8}
 &  & \textbf{Examples} & \textbf{Tokens} & \textbf{Examples} & \textbf{Tokens} & \textbf{Examples} & \textbf{Tokens} \\
\hline
\multirow{2}{*}{\emph{AfriHG}} & Xho & 10,440 & 5,892,814 & 1,305 & 750,506 & 1,305 & 734,845 \\
 & Zul & 14,209 & 7,495,625 & 1,777 & 952,525 & 1,776 & 944,750 \\
\hline
\multirow{1}{*}{\emph{T2X}} & Xho & 3,859 & 346,518 & 460 & 44,208 & 378 & 44,296 \\
\hline
\multirow{4}{*}{\emph{INJONGO Intent}} & Eng & 1,779 & 398,170 & 622 & 139,312 & 622 & 139,312 \\
 & Sot & 2,240 & 501,824 & 320 & 71,720 & 640 & 143,370 \\
 & Xho & 2,240 & 509,035 & 320 & 72,795 & 640 & 145,383 \\
 & Zul & 2,240 & 508,096 & 320 & 72,624 & 640 & 145,139 \\
\hline
\multirow{3}{*}{\emph{MasakhaNER~2.0}} & Tsn & 3,489 & 609,722 & 499 & 87,620 & 996 & 173,044 \\
 & Xho & 5,718 & 920,599 & 817 & 142,658 & 1,633 & 274,254 \\
 & Zul & 5,848 & 922,761 & 836 & 134,701 & 1,670 & 266,979 \\
\hline
\multirow{2}{*}{\emph{MasakhaNEWS}} & Eng & 3,309 & 2,594,818 & 472 & 369,539 & 948 & 752,019 \\
 & Xho & 1,032 & 607,254 & 147 & 84,010 & 297 & 173,229 \\
\hline
\multirow{6}{*}{\emph{SIB-200}} & Afr & 701 & 78,995 & 99 & 10,851 & 204 & 22,691 \\
 & Eng & 701 & 77,191 & 99 & 10,635 & 204 & 22,267 \\
 & Nso & 701 & 91,974 & 99 & 12,468 & 204 & 26,812 \\
 & Sot & 701 & 87,709 & 99 & 12,064 & 204 & 25,394 \\
 & Xho & 701 & 81,956 & 99 & 11,281 & 204 & 23,645 \\
 & Zul & 701 & 81,336 & 99 & 11,198 & 204 & 23,429 \\
\hline
\multirow{3}{*}{\emph{MasakhaPOS}} & Tsn & 754 & 462,757 & 150 & 88,279 & 602 & 342,468 \\
 & Xho & 752 & 364,051 & 150 & 69,079 & 601 & 281,026 \\
 & Zul & 753 & 349,628 & 150 & 68,687 & 601 & 269,956 \\
\hline
\textbf{TOTAL} &  --  & \textbf{62,868} & \textbf{22,982,833} & \textbf{8,939} & \textbf{3,216,760} & \textbf{14,573} & \textbf{4,974,308} \\
\hline
\end{tabular}
\end{adjustbox}
\caption{Supervised data used for instruction tuning, reported per task and language. Counts include both prompt and completion tokens.}
\label{tab:finetune_by_task_lang}
\end{table*}

\begin{table}[t]
\centering
\small
\setlength{\tabcolsep}{4pt}
\begin{adjustbox}{center,max width=\textwidth}
\begin{tabular}{l c c c c}
\hline
\textbf{Model / Variant} & \textbf{Size} & \textbf{BLEU} & \textbf{chrF} & \textbf{ROUGE-L} \\
\hline
\multicolumn{5}{l}{\textbf{T2X (isiXhosa)}}\\
\hline
\textbf{MzansiLM} &  &  &  &  \\
\subsetting{Base 0-shot}       & 0.125B & 0.00 & 0.03 & 3.29 \\
\subsetting{Base 1-shot}       & 0.125B & 0.00 & 0.03 & 4.08 \\
\subsetting{Base 3-shot}       & 0.125B & 0.00 & 0.00 & 0.74 \\
\subsetting{mono-t2x-ft}       & 0.125B & \textbf{20.65} & \textbf{31.56} & \textbf{41.19} \\
\subsetting{general-ft}        & 0.125B & 0.00 & 0.00 & 2.05 \\
\hdashline
\multicolumn{5}{l}{\textit{Encoder--Decoder}} \\
\quad mT5-base          & 0.58B  & \underline{16.8} & \underline{28.7} & \underline{38.7} \\
\quad Aya               & 13B    & 8.9 & 22.1 & 33.9 \\
\hline
\multicolumn{5}{l}{\textbf{AfriHG (isiXhosa)}}\\
\hline
\textbf{MzansiLM} &  &  &  &  \\
\subsetting{Base 0-shot}       & 0.125B & 0.00 & 0.03 & 3.29 \\
\subsetting{Base 1-shot}       & 0.125B & 0.00 & 0.03 & 4.08 \\
\subsetting{Base 3-shot}       & 0.125B & 0.00 & 0.00 & 0.74 \\
\subsetting{mono-afrihg-ft}    & 0.125B & \underline{0.63} & 4.47 & \underline{14.28} \\
\subsetting{multi-afrihg-ft}   & 0.125B & \textbf{0.95} & 5.67 & \textbf{16.98} \\
\subsetting{general-ft}        & 0.125B & 0.36 & 4.10 & 12.42 \\
\hdashline
\multicolumn{5}{l}{\textit{Encoder--Decoder}} \\
\quad AfriTeVa2 base  & 0.313B & -- & \underline{14.9} & -- \\
\quad mT5-base          & 0.58B  & -- & 12.7 & -- \\
\quad Aya               & 13B    & -- & \textbf{15.2} & -- \\
\hline
\multicolumn{5}{l}{\textbf{AfriHG (isiZulu)}}\\
\hline
\textbf{MzansiLM} &  &  &  &  \\
\subsetting{Base 0-shot}       & 0.125B & 0.00 & 0.07 & 4.24 \\
\subsetting{Base 1-shot}       & 0.125B & 0.00 & 0.07 & 4.61 \\
\subsetting{Base 3-shot}       & 0.125B & 0.00 & 0.29 & 0.86 \\
\subsetting{mono-afrihg-ft}    & 0.125B & \textbf{1.33} & 9.67 & \textbf{19.05} \\
\subsetting{multi-afrihg-ft}   & 0.125B & \underline{0.79} & 7.05 & \underline{17.48} \\
\subsetting{general-ft}        & 0.125B & 0.00 & 3.34 & 10.08 \\
\hdashline
\multicolumn{5}{l}{\textit{Encoder--Decoder}} \\
\quad AfriTeVa2 base  & 0.313B & -- & \textbf{17.4} & -- \\
\quad mT5-base          & 0.58B  & -- & 15.5 & -- \\
\quad Aya               & 13B    & -- & \underline{16.2} & -- \\
\hline
\end{tabular}
\end{adjustbox}
\caption{
Combined results for isiXhosa and isiZulu across T2X (data-to-text) and AfriHG (headline generation). 
Boldface indicates best performance; underline indicates second best.
}
\label{tab:combined_gen}
\end{table}

\begin{table}[t]
\centering
\small
\setlength{\tabcolsep}{4pt}
\begin{tabular}{l c c c}
\hline
\textbf{Model / Variant} & \textbf{Size} & \textbf{Eng} & \textbf{Xho} \\
\hline
\textbf{MzansiLM}                     &         &        &        \\
\subsetting{Base 0-shot}               & 0.125B  & 38.3  & 39.1  \\
\subsetting{mono-masakhanews-ft}              & 0.125B  & 63.5  & 73.2  \\
\subsetting{multi-masakhanews-ft}              & 0.125B  & 60.8  & 78.5  \\
\subsetting{general-ft}                        & 0.125B  & 49.2  & 50.9  \\
\hdashline
\multicolumn{4}{l}{\textit{Decoder--Only}} \\
InkubaLM-0.4B                     & 0.4B    & 20.3  & 7.4  \\
AfroLlama-V1                      & 8B      & 67.1  & 50.2  \\
Llama-3.1-70B-Instruct            & 70B     & 83.3  & 68.4  \\
\hdashline
\multicolumn{4}{l}{\textit{Encoder--Decoder}} \\
\quad Aya-101                           & 13B     & 87.1  & 94.6  \\
\hdashline
\multicolumn{4}{l}{\textit{Encoder--Only}} \\
AfriBERTa                         & 0.126B  & 88.9  & 87.0  \\
AfroXLMR-base                     & 0.270B  & \underline{92.2} &   \underline{94.7} \\
AfroXLMR-large                    & 0.550B  & \textbf{93.1} & \textbf{97.3} \\
\hline
\end{tabular}
\caption{Macro-F1 scores for MasakhaNEWS topic classification. Boldface marks the best performance, while underline marks the second best.}
\label{tab:masakhanews}
\end{table}

\begin{table}[t]
\centering
\small
\setlength{\tabcolsep}{4pt}
\begin{tabular}{l c c c c}
\hline
\textbf{Model / Variant} & \textbf{Size} & \textbf{Xho} & \textbf{Zul} & \textbf{Tsn} \\
\hline
\textbf{MzansiLM}                     &         &        &        &        \\
\subsetting{Base 0-shot}               & 0.125B  & 0.0  & 0.0  & 0.0  \\
\subsetting{mono-masakhaner-ft }              & 0.125B  & 48.4  & 26.1  & 38.9  \\
\subsetting{multi-masakhaner-ft}               & 0.125B  & 24.8  & 22.6  & 23.7  \\
\subsetting{general-ft}                        & 0.125B  & 13.6  & 20.7  & 12.3  \\
\hdashline
\multicolumn{5}{l}{\textit{Decoder--Only}} \\
\quad InkubaLM-0.4B                     & 0.4B    & 0.1  & 0.0  & 0.0  \\
\quad AfroLlama-V1                      & 8B      & 5.6  & 3.8  & 3.4  \\
\quad Llama-3.1-70B-Instruct            & 70B     & 8.2  & 10.1  & 20.7  \\
\hdashline
\multicolumn{5}{l}{\textit{Encoder--Decoder}} \\
\quad Aya-101                           & 13B     & 0.0  & 0.0  & 0.0  \\
\hdashline
\multicolumn{5}{l}{\textit{Encoder--Only}} \\
\quad AfriBERTa                         & 0.126B  & 85.0  & 81.7  & 83.2  \\
\quad AfroXLMR-base                     & 0.270B  & \underline{88.6} & \underline{88.4} & \underline{87.7} \\
\quad AfroXLMR-large                    & 0.550B  & \textbf{89.9} & \textbf{90.6} & \textbf{89.4} \\
\hline
\end{tabular}
\caption{Macro-F1 scores for MasakhaNER~2.0 named entity recognition. Boldface marks the best performance, while underline marks the second best.}
\label{tab:masakhaner}
\end{table}

\begin{table}[t]
\centering
\small
\setlength{\tabcolsep}{4pt}
\begin{tabular}{l c c c c}
\hline
\textbf{Model / Variant} & \textbf{Size} & \textbf{Xho} & \textbf{Zul} & \textbf{Tsn} \\
\hline
\textbf{MzansiLM} & & & & \\
\subsetting{Base 0-shot} & 0.125B & 0.0 & 0.0 & 0.0 \\
\subsetting{mono-masakhapos-ft} & 0.125B & 30.5 & 37.8 & 5.6 \\
\subsetting{multi-masakhapos-ft} & 0.125B & 11.6 & 9.5 & 0.5 \\
\subsetting{general-ft} & 0.125B & 0.0 & 0.0 & 0.0 \\
\hdashline
\multicolumn{5}{l}{\textit{Decoder--Only}} \\
\quad InkubaLM-0.4B & 0.4B & 0.0 & 0.0 & 0.0 \\
\quad AfroLlama-V1 & 8B & 0.0 & 0.0 & 0.0 \\
\quad Llama-3.1-8B-Instruct & 8B & 58.1 & 64.2 & 49.5 \\
\quad Llama-3.1-70B-Instruct & 70B & 68.6 & 71.6 & 52.2 \\
\hdashline
\multicolumn{5}{l}{\textit{Encoder--Decoder}} \\
\quad Aya-101 & 13B & 0.0 & 0.0 & 0.0 \\
\hdashline
\multicolumn{5}{l}{\textit{Encoder--Only}} \\
\quad AfriBERTa & 0.126B & 86.1 & 86.9 & 82.5 \\
\quad AfroXLMR-base & 0.270B & \underline{88.5} & \underline{89.4} & \underline{82.7} \\
\quad AfroXLMR-large & 0.550B & \textbf{88.7} & \textbf{90.1} & \textbf{83.0} \\
\hline
\end{tabular}
\caption{Token accuracy for MasakhaPOS part-of-speech tagging. Boldface marks the best performance, while underline marks the second best.}
\label{tab:masakhapos}
\end{table}

\begin{table*}[t]
\centering
\small
\setlength{\tabcolsep}{4pt}
\begin{adjustbox}{center,max width=\textwidth}
\begin{tabular}{c l c c c c c c c}
\hline
\textbf{Task} & \textbf{Model / Variant} & \textbf{Size} & \textbf{Eng} & \textbf{Xho} & \textbf{Zul} & \textbf{Sot} & \textbf{Nso} & \textbf{Afr} \\
\hline
\multirow{13}{*}{\emph{SIB-200}} 
  & \textbf{MzansiLM}                     &         &        &        &        &       &       &       \\
  & \subsetting{Base 0-shot}               & 0.125B  & 32.3  & 28.9  & 31.2  & 18.0 & 17.3 & 43.4 \\
  & \subsetting{mono-sib200-ft}                    & 0.125B  & 28.0  & 39.1  & 22.0  & 36.8 & 36.6 & 54.4 \\
  & \subsetting{multi-sib200-ft}                   & 0.125B  & 33.3  & 40.4  & 47.2  & 34.7 & 30.5 & 29.6 \\
  & \subsetting{general-ft}                        & 0.125B  & 28.0  & 39.1  & 47.2  & 36.8 & 33.9 & 54.4 \\
\hdashline
  & \multicolumn{8}{l}{\textit{Decoder--Only}} \\
  & \quad InkubaLM-0.4B                     & 0.4B    & 9.0  & 8.4  & 8.2  & 5.3 & 6.4 & 5.3 \\
  & \quad AfroLlama-V1                      & 8B      & 6.4  & 6.5  & 6.4  & 39.7 & 38.7 & 6.4 \\
  & \quad Llama-3.1-70B-Instruct            & 70B     & 88.3  & 65.0  & 57.3  & 54.4 & 55.8 & 85.6 \\
\hdashline
  & \multicolumn{8}{l}{\textit{Encoder--Decoder}} \\
  & \quad Aya-101                           & 13B     & 82.8  & 82.0  & 82.9  & 81.4 & 82.1 & 83.7 \\
\hdashline
  & \multicolumn{8}{l}{\textit{Encoder--Only}} \\
  & \quad AfriBERTa                         & 0.126B  & --     & 70.7  & 73.5  & 55.9 & 54.8 & 89.8 \\
  & \quad AfroXLMR-base                     & 0.270B  & --     & \underline{83.1} & \underline{84.9} & \textbf{83.7} & \underline{80.7} & \underline{90.4} \\
  & \quad AfroXLMR-large                    & 0.550B  & --     & \textbf{84.0} & \textbf{85.8} & \underline{83.5} & \textbf{83.3} & \textbf{91.1} \\
\hline

\multirow{12}{*}{\emph{INJONGO Intent}}
  & \textbf{MzansiLM}                     &         &        &        &        &       &       &       \\
  & \subsetting{Base 0-shot}               & 0.125B  & 3.1  & 0.8  & 0.9  & 1.4 & --    & --    \\
  & \subsetting{mono-injongo-intent-ft}            & 0.125B  & 3.4  & 0.6  & 0.6  & 0.9 & --    & --    \\
  & \subsetting{multi-injongo-intent-ft}           & 0.125B  & 2.9  & 1.0  & 0.7  & 0.7 & --    & --    \\
  & \subsetting{general-ft}                        & 0.125B  & 3.5  & 0.6  & 0.6  & 0.9 & --    & --    \\
\hdashline
  & \multicolumn{8}{l}{\textit{Decoder--Only}} \\
  & \quad InkubaLM-0.4B                     & 0.4B    & 0.4  & 0.3  & 0.3  & 0.3 & --    & --    \\
  & \quad AfroLlama-V1                      & 8B      & 10.1  & 1.8  & 1.0  & 0.1 & --    & --    \\
  & \quad Llama-3.1-70B-Instruct            & 70B     & 84.7  & 34.3  & 35.2  & 23.3 & --    & --    \\
\hdashline
  & \multicolumn{8}{l}{\textit{Encoder--Decoder}} \\
  & \quad Aya-101                           & 13B     & 70.7  & 62.7  & 57.2  & 45.9 & --    & --    \\
\hdashline
  & \multicolumn{8}{l}{\textit{Encoder--Only}} \\
  & \quad AfriBERTa                         & 0.126B  & 74.2  & 94.4  & 95.0  & 81.9 & -- & -- \\
  & \quad AfroXLMR-base                     & 0.270B  & \underline{84.1} & \textbf{97.3} & \underline{96.1} & \underline{85.5} & -- & -- \\
  & \quad AfroXLMR-large                    & 0.550B  & \textbf{84.5} & \underline{96.9} & \textbf{97.7} & \textbf{86.8} & -- & -- \\
\hline
\end{tabular}
\end{adjustbox}
\caption{Accuracy for SIB-200 (topic classification) and INJONGO Intent (intent classification). Boldface marks the best performance, while underline marks the second best.
}
\label{tab:acc_tasks_only}
\end{table*}

\begin{table*}[t]
\centering
\small
\setlength{\tabcolsep}{4pt}
\begin{adjustbox}{center,max width=\textwidth}
\begin{tabular}{l c c c c c c c c}
\hline
\textbf{Model / Variant} & \textbf{Size} & \textbf{Xho} & \textbf{Zul} & \textbf{Tsn} & \textbf{Ssw} & \textbf{Sot} & \textbf{Eng} & \textbf{Afr} \\
\hline
\textbf{MzansiLM}                 &        &        &        &        &        &        &        &        \\
\subsetting{Base 0-shot}           & 0.125B & 27.8  & 28.0  & 28.3  & 27.8  & 27.1  & 27.3  & 27.3 \\
\subsetting{Base 1-shot}           & 0.125B & 29.4  & 28.2  & 30.8  & 29.7  & 27.4  & 27.4  & 27.6 \\
\subsetting{Base 3-shot}           & 0.125B & 28.8  & 28.2  & 29.0  & 28.8  & 27.3  & 27.6  & 27.4 \\
\subsetting{general-ft}                    & 0.125B & 21.9  & 25.1  & 21.9  & 23.6  & 22.0  & 29.6  & 31.1 \\
\hdashline
\multicolumn{9}{l}{\textit{Decoder--Only}}\\
\quad InkubaLM-0.4B                 & 0.4B   & 23.1  & 23.2  & 24.6  & --     & --     & 23.9  & 25.9 \\
\quad AfroLlama-V1                  & 8B     & 24.8  & 28.2  & 26.8  & 22.9  & 22.6  & 25.3  & 24.7 \\
\quad Llama-3.1-8B-Instruct         & 8B     & 35.1  & 35.3  & 32.3  & 32.3  & 33.7  & 80.7  & 66.9 \\
\quad Meta-Llama-3-70B-Instruct     & 70B    & 41.3  & 42.9  & \underline{41.4}  & 42.9  & \underline{51.3}  & \textbf{93.2}  & \textbf{88.9} \\
\hdashline
\multicolumn{9}{l}{\textit{Encoder--Decoder}}\\
\quad Aya-101                       & 13B    & \textbf{65.9} & \textbf{64.9} & \textbf{63.6} & \textbf{57.6} & \textbf{61.7} & \underline{86.1}  & \underline{81.7} \\
\hdashline
\multicolumn{9}{l}{\textit{Encoder--Only}}\\
\quad XLM-V large                   & --     & \underline{54.4} & \underline{54.2} & --     & \underline{47.1} & 32.7 & 77.8  & 72.3 \\
\hline
\end{tabular}
\end{adjustbox}
\caption{Normalised accuracy for Belebele reading comprehension. Boldface marks the best performance, while underline marks the second best. 
}
\label{tab:belebele}
\end{table*}

\begin{table*}[t]
\centering
\small
\setlength{\tabcolsep}{4pt}
\begin{adjustbox}{center,max width=0.95\textwidth}
\begin{tabular}{l c @{\hspace{8pt}} c c c @{\hspace{8pt}} c c c c @{\hspace{8pt}} c c c c}
\hline
 & & \multicolumn{3}{c@{\hspace{8pt}}}{\textbf{AfriXNLI}} & \multicolumn{4}{c@{\hspace{8pt}}}{\textbf{AfriMMLU}} & \multicolumn{4}{c}{\textbf{AfriMGSM}} \\
\hline
\textbf{Model / Variant} & \textbf{Size} & \textbf{Xho} & \textbf{Zul} & \textbf{Sot} & \textbf{Xho} & \textbf{Zul} & \textbf{Sot} & \textbf{Eng} & \textbf{Xho} & \textbf{Zul} & \textbf{Sot} & \textbf{Eng} \\
\hline
\textbf{MzansiLM} & & & & & & & & & & & & \\
\subsetting{Base 0-shot} & 0.125B & 32.3 & 32.0 & 32.3 & 22.7 & 24.7 & 23.7 & 25.4 & 1.6 & 2.4 & 1.2 & 0.8 \\
\subsetting{general-ft} & 0.125B & 33.4 & 33.6 & 31.8 & 23.0 & 24.2 & 25.1 & 23.4 & 2.4 & 0.8 & 1.6 & 1.6 \\
\hdashline
\multicolumn{13}{l}{\textit{Decoder--Only}} \\
\quad InkubaLM-0.4B & 0.4B & 34.3 & 34.2 & 33.5 & 26.2 & 23.4 & 26.6 & 24.3 & \underline{9.2} & \underline{9.2} & 2.4 & 12.0 \\
\quad AfroLlama-V1 & 8B & \underline{41.8} & \underline{38.3} & 34.3 & 26.6 & 27.1 & 26.0 & 34.0 & 0.0 & 0.4 & 0.8 & 0.4 \\
\quad Llama-3.1-8B-Instruct & 8B & 34.0 & 34.7 & 34.8 & 27.2 & \underline{31.8} & 31.8 & \underline{62.8} & 4.4 & 5.2 & 6.4 & \underline{56.8} \\
\quad Llama-3.1-70B-Instruct & 70B & 39.3 & 37.0 & \underline{39.0} & \textbf{34.2} & \textbf{40.4} & \textbf{39.0} & \textbf{76.4} & \textbf{18.8} & \textbf{27.2} & \textbf{32.8} & \textbf{86.8} \\
\hdashline
\multicolumn{13}{l}{\textit{Encoder--Decoder}} \\
\quad Aya-101 & 13B & \textbf{55.2} & \textbf{55.3} & \textbf{54.7} & \underline{32.2} & 31.2 & \underline{33.2} & 42.8 & 4.4 & 4.8 & \underline{10.8} & 11.6 \\
\hline
\end{tabular}
\end{adjustbox}
\caption{Accuracy for IrokoBench instruction following tasks: AfriXNLI, AfriMMLU, and AfriMGSM. 
Boldface marks the best performance, while underline marks the second best.
}
\label{tab:acc_tasks_cols}
\end{table*}

\section{Instruction Finetuning}

We adapt the base MzansiLM to several tasks. In our experiments we test three adaptation regimes that differ in their level of generalization: \emph{monolingual task-specific finetuning} represents the narrowest scope, \emph{multilingual task-specific finetuning} pools related languages under a shared task structure, and \emph{multi-task instruction finetuning} combines all tasks and languages in a single training mixture. 
It remains unclear from the current literature which regime is most effective at this model scale and how this differs across languages and tasks. Prior work shows that the effectiveness of monolingual versus multilingual finetuning varies by task and language pair \citep{eisenschlos-etal-2019-multifit,tang-etal-2021-multilingual}, while multi-task instruction tuning introduces additional tradeoffs between task coverage and task-specific performance \citep{uemura-etal-2024-afriinstruct}.

\subsection{Datasets}
\label{subsec:datasets}

We identified Natural Language Understanding and Generation datasets covering multiple South African languages which are suited for instruction finetuning. 
Table~\ref{tab:finetune_by_task_lang} reports train, validation, and test counts for the supervised datasets in our main instruction-tuning mixture. We use existing splits when provided; otherwise, we create a stratified validation split from the training data for early stopping. 

\paragraph{Document classification}
MasakhaNEWS \citep{adelani-etal-2023-masakhanews} and SIB-200 \citep{adelani-etal-2024-sib} provide document-level topic classification. MasakhaNEWS contains news articles across seven news categories 
while SIB-200 contains sentences labelled into seven categories. 
INJONGO Intent \citep{yu-etal-2025-injongo} targets intent classification with 40 intent classes across five domains.

\paragraph{Sequence labelling}
For sequence labelling, we use MasakhaPOS \citep{dione-etal-2023-masakhapos} for part-of-speech tagging and MasakhaNER~2.0 \mbox{\citep{adelani-etal-2022-masakhaner}} for named entity recognition.

\paragraph{Generation}
AfriHG \citep{ogunremi2024afrihg} evaluates summarization via headline generation for isiXhosa and isiZulu, while T2X \citep{meyer-buys-2024-triples} assesses data-to-text generation (mapping triples to descriptive sentences) in isiXhosa.
The availability of suitable generation datasets for most of the target languages remain very limited. 

\paragraph{Reading comprehension and few-shot understanding}
Belebele  \citep{bandarkar-etal-2024-belebele} evaluates multiple-choice reading comprehension. 
IrokoBench \citep{adelani-etal-2025-irokobench} is a benchmark suite for evaluating few-shot reasoning on African languages, comprising three tasks: AfriXNLI for natural language inference, AfriMMLU for knowledge-based multiple-choice questions,
and AfriMGSM for mathematical word problems. 
These tasks are challenging for even large-scale LMs, many of which exhibit chance performance \citep{adelani-etal-2025-irokobench}.

\subsection{Finetuning Setup}

We format all supervised tasks as next-token prediction, preserving the decoder-only architecture without introducing encoder components or task-specific heads. Inputs use a maximum length of 2{,}048 tokens for classification and labelling and 1{,}024 tokens for generation. We apply early stopping on each task's validation split and select checkpoints by the task's primary validation metric, using token accuracy for sequence labelling and task-standard metrics elsewhere. When multiple instruction templates are available, training cycles through them to cover the formats encountered in zero-, one-, and three-shot prompting.

For most tasks, we use the prompt templates from LM Evaluation Harness \citep{eval-harness} to ensure consistency between training and evaluation formats. For T2X and AfriHG, we use standard templates that prompt the model with an input and request the corresponding output.

We employ LoRA for parameter-efficient finetuning with rank 128, scaling $\alpha=256$, and dropout 0.1, applied to attention and feed-forward projections. Most tasks have limited training data and models overfit without parameter-efficient adaptation such as LoRA. We conducted Bayesian hyperparameter tuning and use settings adapted from pretraining unless a task requires a shorter training horizon. We set the number of finetuning epochs for each task: MasakhaNEWS, INJONGO Intent, and MasakhaPOS train for 20 epochs, SIB-200 for 15 epochs, MasakhaNER~2.0 for 50 epochs, AfriHG for 3 epochs, and T2X for 4 epochs.

We compare the following adaptation approaches of our base MzansiLM:
\begin{itemize}
    \item \textbf{Base 0-shot:} no finetuning and no in-context examples at evaluation.
    \item \textbf{Base few-shot:} no finetuning but 1 or 3 in-context examples at evaluation.
    \item \textbf{Monolingual task-specific finetuning:} finetune a separate model for each task–language pair using only examples from that language and task.
    \item \textbf{Multilingual task-specific finetuning:} finetune one model per task by pooling all languages with examples for that task, enabling cross-lingual transfer in a fixed label space.
    \item \textbf{Multi-task instruction finetuning:} finetune a single model on all tasks and languages jointly using a weighted sampling mixture.
\end{itemize}

In the multi-task regime we use fixed, per-dataset sampling weights to balance supervision across the task types. 
Across all regimes, we reuse the tokenizer described in Section~\ref{sec:pretrain}.
Test-time decoding follows task-appropriate settings. Results in Tables~\ref{tab:combined_gen} through \ref{tab:acc_tasks_cols} come directly from the three regimes defined here.


\section{Results}
\label{sec:results}

We evaluate MzansiLM and its finetuned variants across all tasks listed in Section~\ref{subsec:datasets}. The results are presented in Tables~\ref{tab:combined_gen}--\ref{tab:acc_tasks_cols}. This section addresses two research questions. First, how do the three finetuning strategies—monolingual task-specific, multilingual task-specific, and general multi-task instruction—compare to each other? Second, how competitive is MzansiLM at the 125M scale against larger baseline models?

\subsection{Comparing Finetuning Strategies}

All finetuning regimes raise performance well above base model prompting
for document classification. 
On MasakhaNEWS (Table~\ref{tab:masakhanews}), multilingual task-specific finetuning achieves the highest macro-F1 in isiXhosa, benefiting from shared signal across closely related Bantu languages, while monolingual task-specific finetuning is competitive in English. On SIB-200 (Table~\ref{tab:acc_tasks_only}), both monolingual and multilingual task-specific finetuning improve accuracy relative to base prompting, but the leading variant varies by language and no single regime dominates. INJONGO Intent performance (Table~\ref{tab:acc_tasks_only}) remains near chance (2.5\%) across all finetuning regimes at this model size.
For sequence labelling (Tables~\ref{tab:masakhaner} and \ref{tab:masakhapos}), base model prompting does not produce usable output. Monolingual task-specific finetuning is the most effective adaptation regime, while multilingual task-specific finetuning and general multi-task finetuning are weaker at this model scale.

Monolingual task-specific finetuning performs decisively stronger than base prompting and general multi-task finetuning on isiXhosa data-to-text generation (Table~\ref{tab:combined_gen}), with the gains indicating that targeted supervision is essential at this scale. 
Headline generation in isiXhosa and isiZulu remains challenging, with base model prompting near zero. Among the finetuned variants, multilingual task-specific finetuning is strongest for isiXhosa while monolingual task-specific finetuning leads in isiZulu (Table~\ref{tab:combined_gen}).

Normalized accuracy on Belebele reading comprehension (Table~\ref{tab:belebele})  clusters near chance for South African languages under base prompting. General multi-task finetuning improves accuracy on English and Afrikaans but not the target languages. 
On the IrokoBench evaluation suite for few-shot reasoning (Table~\ref{tab:acc_tasks_cols}) general multi-task finetuning does not improve performance over base prompting at this scale.

\subsection{Comparison to Baseline Models}

MzansiLM demonstrates competitive performance against larger models on several specific tasks. 
Monolingual task-specific finetuning on T2X data-to-text generation reaches 20.65 BLEU on isiXhosa, surpassing mT5-base and competing with encoder--decoder baselines over ten times larger (Table~\ref{tab:combined_gen}). 
On named entity recognition, monolingual task-specific finetuning substantially outperforms all other decoder-only baselines, including InkubaLM at 0.4B parameters, AfroLlama-V1 at 8B parameters, and even Llama-3.1-70B-Instruct, demonstrating that task-specific supervision at small scale can exceed few-shot prompting of much larger models (Table~\ref{tab:masakhaner}). 
Multilingual task-specific finetuning achieves 78.5\% macro-F1 in isiXhosa on MasakhaNEWS topic classification, outperforming both InkubaLM and AfroLlama-V1 (Table~\ref{tab:masakhanews}). 
However, MzansiLM remains less competitive on part-of-speech tagging and named entity recognition, with meaningful gains only from monolingual task-specific finetuning and substantially stronger performance from larger prompted or encoder-based baselines (Tables~\ref{tab:masakhapos} and \ref{tab:masakhaner}).

On tasks where MzansiLM achieves near-chance performance, it matches or exceeds other decoder-only models at comparable or larger scales, indicating that these tasks are fundamentally challenging for decoder-only architectures below 70B parameters. On INJONGO Intent, MzansiLM's performance is comparable to InkubaLM at 0.4B and AfroLlama-V1 at 8B, with only Llama-3.1-70B-Instruct showing substantial improvement (Table~\ref{tab:acc_tasks_only}). Similarly, MzansiLM's normalized accuracy
on Belebele reading comprehension matches InkubaLM and AfroLlama-V1 across South African languages, with all decoder-only models below 70B parameters remaining near chance (Table~\ref{tab:belebele}). On AfriXNLI, AfriMMLU, and AfriMGSM, the pattern repeats: MzansiLM performance aligns with decoder-only models up to 8B parameters, and substantial gains require scaling to 70B (Table~\ref{tab:acc_tasks_cols}).

Encoder-only models retain a clear advantage on high-accuracy classification and sequence labelling tasks. AfroXLMR-large consistently achieves macro-F1 scores above 89 on both MasakhaNEWS and MasakhaNER~2.0, substantially ahead of MzansiLM's finetuned variants (Tables~\ref{tab:masakhanews} and \ref{tab:masakhaner}). On SIB-200 and INJONGO Intent, encoder-only baselines reach accuracy in the 80-90 range while all decoder-only models remain far below, with INJONGO Intent particularly challenging for decoder-only architectures at any scale below 70B (Table~\ref{tab:acc_tasks_only}). Encoder–decoder models dominate reading comprehension and headline generation. Aya-101 at 13B parameters achieves normalized accuracy of 65.9 on Belebele isiXhosa, far exceeding all decoder-only models including those at 70B scale (Table~\ref{tab:belebele}). Similarly, AfriTeVa V2 and Aya-101 substantially outperform MzansiLM on AfriHG headline generation (Table~\ref{tab:combined_gen}).

Overall, these results show that task-specific supervision is most effective for generation and sequence labelling at 125M parameters, that multilingual task-specific finetuning can help closely related Bantu languages on topic classification, and that encoder-based models remain preferable for more structured prediction tasks. Few-shot multiple-choice reasoning, reading comprehension, and intent classification are difficult for decoder-only models below 70B parameters, indicating that these tasks require further scaling. For some languages, this is not possible, in which case encoder-only and encoder--decoder architectures remain the more viable and data-efficient choice.



  
\section{Conclusion}


This paper introduced MzansiText and MzansiLM, establishing an open, reproducible decoder-only LM for all eleven written South African languages. Evaluations across three adaptation regimes show that task-specific supervision is most effective at this scale, while few-shot reasoning and reading comprehension remain challenging below 70B parameters.
Our findings clarify the role of small-scale decoder-only LMs for low-resource languages. Rather than replacing encoder-only and encoder--decoder models on tasks where those architectures remain stronger, such as classification and structured prediction, MzansiLM serves as a reproducible decoder-only baseline that helps identify where task-specific finetuning is effective and where alternative architectures are still preferable. At the same time, our results show that smaller models trained from scratch with sound design choices and region-specific language coverage can still match or even outperform much larger models on structured generation tasks.
MzansiText, MzansiLM, and our comprehensive set of results provide a benchmark for future research on language modelling for South African languages.


\section*{Limitations}

We train a 125M-parameter model, which aligns with scaling law expectations given our pretraining data volume and computational budget. This scale enables reproducibility and rapid iteration but constrains our ability to evaluate emergent capabilities such as few-shot reasoning and long-range dependency modeling. The model's token budget limits prompt design for multi-example inputs, and performance on tasks requiring substantial reasoning capacity remains near chance. These constraints are consistent with the scaling behavior observed across decoder-only models at this parameter count and reflect our focus on establishing a reproducible baseline rather than achieving state-of-the-art performance on all tasks.

Data composition is a second limitation. The pretraining mixture is heavily skewed toward Afrikaans and English. As detailed in Section~\ref{sec:dataset}, we experimented with a more balanced corpus but observed measurable degradation on downstream tasks and therefore retained the natural distribution.

While MzansiText provides pretraining coverage for all eleven official languages, our downstream evaluation focuses on the eight languages for which we identified the most directly comparable public benchmark coverage for the tasks considered in this paper. Public evaluation resources do exist for isiNdebele, Tshivenda, and Xitsonga, but their coverage is more fragmented across tasks and does not align as cleanly with the benchmark suite used here.

The general instruction-tuning mixture introduces trade-offs: combining tasks and languages improves coverage but dilutes task-specific signal, particularly for low-resource generative tasks. Evaluation scope is a further constraint. We focus on non-translation tasks and adopt automated metrics, which provide comparability but do not fully capture fluency, error severity, or output quality.


\section*{Ethical Considerations}

All datasets used in this study are publicly available and were collected from open multilingual resources. No private or user-generated data were included. The released model is small in scale, trained for research purposes, and unlikely to cause harm or misuse. We encourage responsible use and clear citation of the datasets and models.

\section*{Acknowledgements}

This work is based on the research supported in part by the National Research Foundation of South Africa (Grant Number 151601) and the Telkom–National Research Foundation (Telkom-NRF) Future Technologies Programme.
Computations were performed using facilities provided by the University of Cape Town’s ICTS High Performance Computing team: \url{https://hpc.uct.ac.za} 



\section{References}\label{sec:reference}
\bibliographystyle{lrec2026-natbib}
\bibliography{lrec2026,languageresource} 

\section{Language Resource References}
\label{lr:ref}
\bibliographystylelanguageresource{lrec2026-natbib}
\bibliographylanguageresource{languageresource}

\appendix

\section{Balanced Pretraining Mixture}
\label{app:balanced_results}

As noted in Section~\ref{sec:dataset}, we also experimented with a more balanced pretraining mixture that reduced the dominance of Afrikaans and English and redistributed more capacity to the nine Bantu languages. We evaluated this balanced-pretraining variant on a subset of downstream tasks to test whether a less skewed corpus would improve adaptation for lower-resource languages.

The results did not support that hypothesis. On MasakhaNEWS and MasakhaNER~2.0, the balanced-pretraining variant generally underperformed the main model trained on the natural data distribution. On SIB-200, INJONGO Intent, Belebele, AfriXNLI, AfriMMLU, and AfriMGSM, performance remained near chance and did not materially differ from the main model. We did not evaluate the balanced-pretraining variant on the generation tasks at that stage, so we do not draw conclusions for AfriHG or T2X.

These exploratory results are consistent with our decision to retain the natural distribution for the main experiments: in this low-resource setting, preserving access to the largest available volume of higher-quality text was more beneficial than enforcing a more balanced language mixture.

\begin{table*}[t]
\centering
\small
\setlength{\tabcolsep}{4pt}
\begin{adjustbox}{center,max width=\textwidth}
\begin{tabular}{l l c c c c c}
\hline
\textbf{Task} & \textbf{Variant} & \textbf{Eng} & \textbf{Xho} & \textbf{Zul} & \textbf{Sot} & \textbf{Tsn} \\
\hline
\multirow{2}{*}{MasakhaNEWS (Macro-F1)}
& mono-ft  & 55.4 & 64.8 & -- & -- & -- \\
& multi-ft & 53.1 & 68.7 & -- & -- & -- \\
\hline
\multirow{2}{*}{MasakhaNER~2.0 (Macro-F1)}
& mono-ft  & -- & 39.2 & 18.7 & -- & 29.8 \\
& multi-ft & -- & 17.4 & 15.9 & -- & 16.8 \\
\hline
\multirow{2}{*}{SIB-200 (Accuracy)}
& Base 0-shot & 31.1 & 27.6 & 29.8 & 17.5 & 16.9 \\
& general-ft  & 27.4 & 36.8 & 44.9 & 34.1 & 29.8 \\
\hline
\multirow{2}{*}{INJONGO Intent (Accuracy)}
& Base 0-shot & 3.0 & 0.7 & 0.8 & 1.2 & -- \\
& general-ft  & 3.2 & 0.6 & 0.6 & 0.8 & -- \\
\hline
\multirow{2}{*}{Belebele (Norm. Acc.)}
& Base 0-shot & 26.9 & 27.1 & 27.4 & 26.8 & 27.7 \\
& general-ft  & 28.7 & 21.4 & 24.5 & 21.7 & 21.0 \\
\hline
\multirow{2}{*}{AfriXNLI (Accuracy)}
& Base 0-shot & -- & 31.9 & 31.5 & 31.7 & -- \\
& general-ft  & -- & 32.6 & 32.8 & 31.1 & -- \\
\hline
\multirow{2}{*}{AfriMMLU (Accuracy)}
& Base 0-shot & 24.8 & 22.1 & 24.0 & 23.1 & -- \\
& general-ft  & 23.0 & 22.6 & 23.7 & 24.4 & -- \\
\hline
\multirow{2}{*}{AfriMGSM (Accuracy)}
& Base 0-shot & 0.8 & 1.2 & 2.0 & 1.2 & -- \\
& general-ft  & 1.2 & 2.0 & 0.8 & 1.2 & -- \\
\hline
\end{tabular}
\end{adjustbox}
\caption{Exploratory downstream results for the balanced-pretraining variant of MzansiLM. The model was evaluated on a subset of downstream tasks only; generation tasks were not included for this variant.}
\label{tab:balanced_appendix_results}
\end{table*}

\end{document}